\documentclass[final]{cvpr}

\pdfoutput=1
\usepackage{times}
\usepackage{epsfig}
\usepackage{graphicx}
\usepackage{amsmath}
\usepackage{amssymb}
\usepackage[linesnumbered,ruled]{algorithm2e}
\usepackage{multirow}
\usepackage{booktabs}
\usepackage{bbm}
\usepackage[accsupp]{axessibility}  

\usepackage[pagebackref=true,breaklinks=true,colorlinks,bookmarks=false]{hyperref}

\def\cvprPaperID{3394} 
\def\confYear{CVPR 2022}

\begin{document}

\title{Domain-Agnostic Prior for Transfer Semantic Segmentation}


\author{
Xinyue Huo\textsuperscript{1,2}\quad Lingxi Xie\textsuperscript{2}\quad Hengtong Hu\textsuperscript{2,3}\quad Wengang Zhou\textsuperscript{1}\quad Houqiang Li\textsuperscript{1}\quad Qi Tian\textsuperscript{2}\\
\textsuperscript{1}University of Science and Technology of China\quad\textsuperscript{2}Huawei Inc.\quad
\textsuperscript{3}Hefei University of Technology\\
\small\texttt{xinyueh@mail.ustc.edu.cn}\quad\small\texttt{198808xc@gmail.com}\quad\small\texttt{huhengtong.hfut@gmail.com}\\\small\texttt{\{zhwg,lihq\}@ustc.edu.cn}\quad\small\texttt{tian.qi1@huawei.com}
}

\maketitle

\begin{abstract}
Unsupervised domain adaptation (UDA) is an important topic in the computer vision community. The key difficulty lies in defining a common property between the source and target domains so that the source-domain features can align with the target-domain semantics. In this paper, we present a simple and effective mechanism that regularizes cross-domain representation learning with a \textbf{domain-agnostic prior} (DAP) that constrains the features extracted from source and target domains to align with a domain-agnostic space. In practice, this is easily implemented as an extra loss term  that requires a little extra costs. In the standard evaluation protocol of transferring synthesized data to real data, we validate the effectiveness of different types of DAP, especially that borrowed from a text embedding model that shows favorable performance beyond the state-of-the-art UDA approaches in terms of segmentation accuracy. Our research reveals that UDA benefits much from better proxies, possibly from other data modalities.
\end{abstract}

\section{Introduction}
\label{introduction}
\begin{figure}[!t]
\centering
\includegraphics[width=0.46\textwidth]{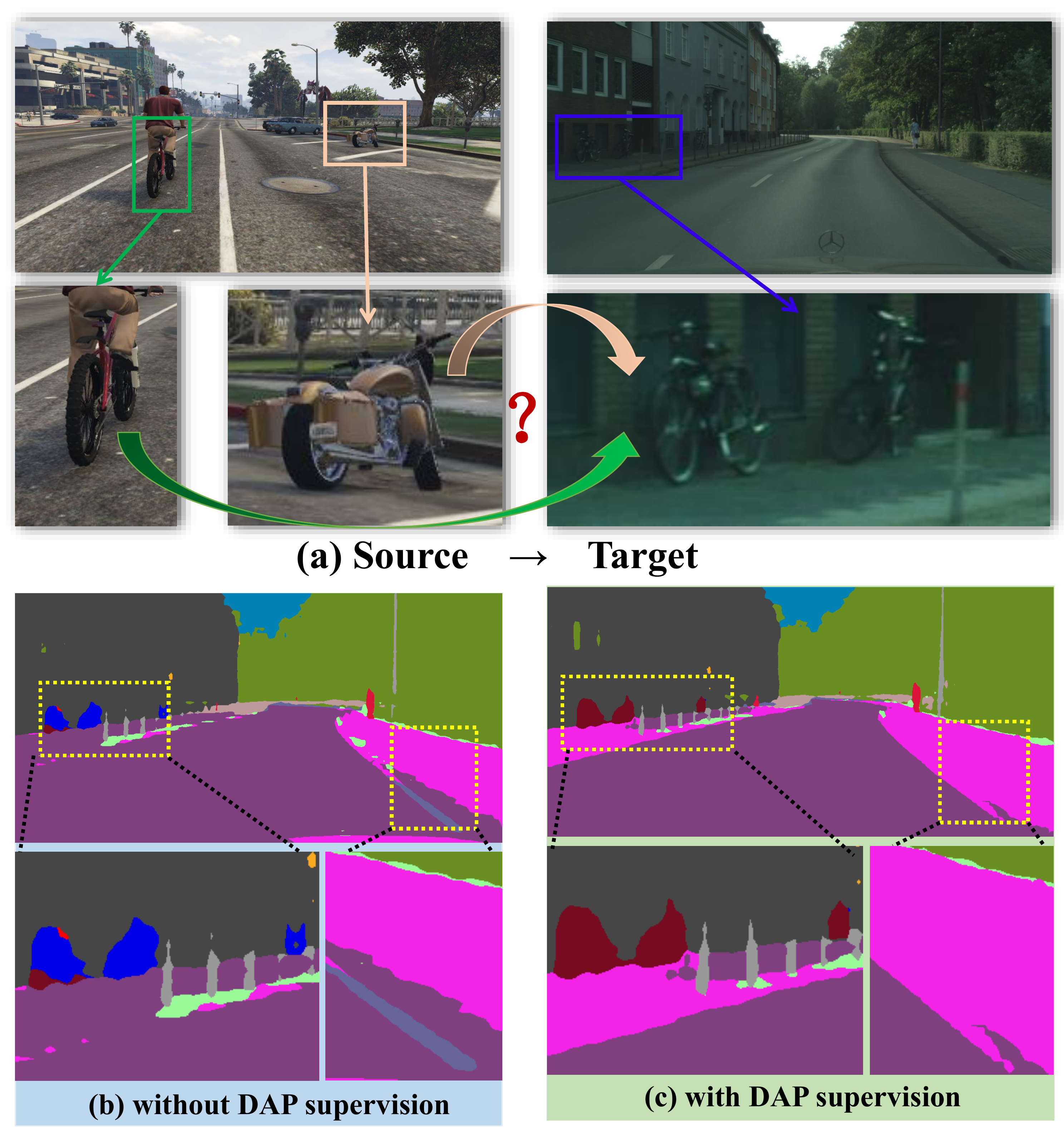}
\caption{\textbf{Top:} The goal is to transfer a segmentation model trained in the source domain to the target domain, but some semantically similar classes (see the left of (a) for examples of \textit{bike} and \textit{motorbike}) are difficult to distinguish due to domain shift. \textbf{Bottom:} segmentation results of the upper-right image without and with DAP, where (b) shows incorrect segmentation (\textit{bike}$\rightarrow$\textit{motorbike} and \textit{sidewalk}$\rightarrow$\textit{road}) of the baseline (DACS~\cite{tranheden2021dacs}), and (c) shows how DAP improves segmentation.}
\label{fig:motivation}
\end{figure}

In the deep learning era, the most powerful approach for visual recognition is to train deep neural networks with abundant, labeled data. Such a data-driven methodology suffers the difficulty of transferring across domains, which raises an important research field named domain adaptation. This paper focuses on the setting of unsupervised domain adaptation (UDA), which assumes that the source domain offers full supervision but the target domain has no annotations available. In particular, we investigate semantic segmentation -- provided the increasing amount of unlabeled data and the expensiveness of annotation, it becomes increasingly important to gain the ability of transferring models from a known domain (\textit{e.g.}, labeled or synthesized data).

The goal of semantic segmentation is to assign each pixel with a class label. In the scenarios that annotations are absent, this is even more challenging than image-level prediction (\textit{e.g.}, classification) because the subtle differences at the pixel level can be easily affected by the change of data distribution. Note that this factor can bring in the risk that pseudo labels are inaccurate and thus deteriorate transfer performance, and the risk becomes even higher when the target domain lacks sufficient data for specific classes or class pairs. Fig~\ref{fig:motivation} shows an example that such approaches~\cite{zou2018unsupervised,tranheden2021dacs} are difficult in distinguishing semantically similar classes (\textit{e.g.}, \textit{motorbike} vs. \textit{bike}, \textit{road} vs. \textit{sidewalk}) from each other.

To alleviate the above issue, we first offer a hypothesis about the confusion -- the proportion of similar categories in the two domains varies too much or they often appear adjacent to each other and the border is difficult to find ( there are limited pixels around the boundary).  Consequently, it is difficult for the deep network to learn the discrimination boundary based on the transferred image features. To compensate, we propose to add a \textbf{domain-agnostic prior} (DAP) to force the features from the source and target domains to align with an auxiliary space that is individual to both domains. According to the Bayesian theory, a properly designed prior relieves the instability of likelihood (provided by the limited training data with similar classes co-occurring) and leads to more accurate inference.

We implement our algorithm upon DACS, a recent approach built upon an advanced data augmentation named ClassMix~\cite{olsson2021classmix}. The training procedure of DACS involves sampling a pair of source and target images and making use of pseudo labels to generate a mixed image with partly-pseudo labels, and feeding the source and mixed images into the deep network. We introduce DAP into the framework by defining a high-dimensional, domain-agnostic embedding vector for each class, and force the features extracted from both the source and mixed images to align with the embedding vectors through an auxiliary module. We investigate two types of domain-agnostic embedding, namely, one-hot vectors and the word2vec features~\cite{mikolov2013distributed}, and show that both of them bring consistent gain in transferring. As a side note, DAP is efficient to carry out -- the auxiliary module is lightweight that requires around $7\%$ extra training computation and is removed during the inference stage.

We evaluate DAP on two standard UDA segmentation benchmarks, in which the source domain is defined by a synthesized dataset (\textit{i.e.}, GTAv~\cite{richter2016playing} or SYNTHIA~\cite{ros2016SYNTHIA}) while the target domain involves Cityscapes~\cite{cordts2016cityscapes}, a dataset captured from the real world. With the word2vec features as prior, DAP achieves segmentation mIOU scores of $55.0\%$ and $50.2\%$ from GTAv and SYNTHIA, respectively, with absolute gains of $2.9\%$ and $1.3\%$ over the DACS baseline, setting the new state-of-the-art among single-model, single-round approaches for UDA segmentation.

The main contribution of this work lies in the proposal and implementation of domain-agnostic prior for UDA segmentation. With such a simple and effective approach, we reveal that much room is left behind UDA. We expect more sophisticated priors and/or more effective constraints to be explored in the future.

\begin{figure*}[!t]
\centering
\includegraphics[width=1.0\textwidth]{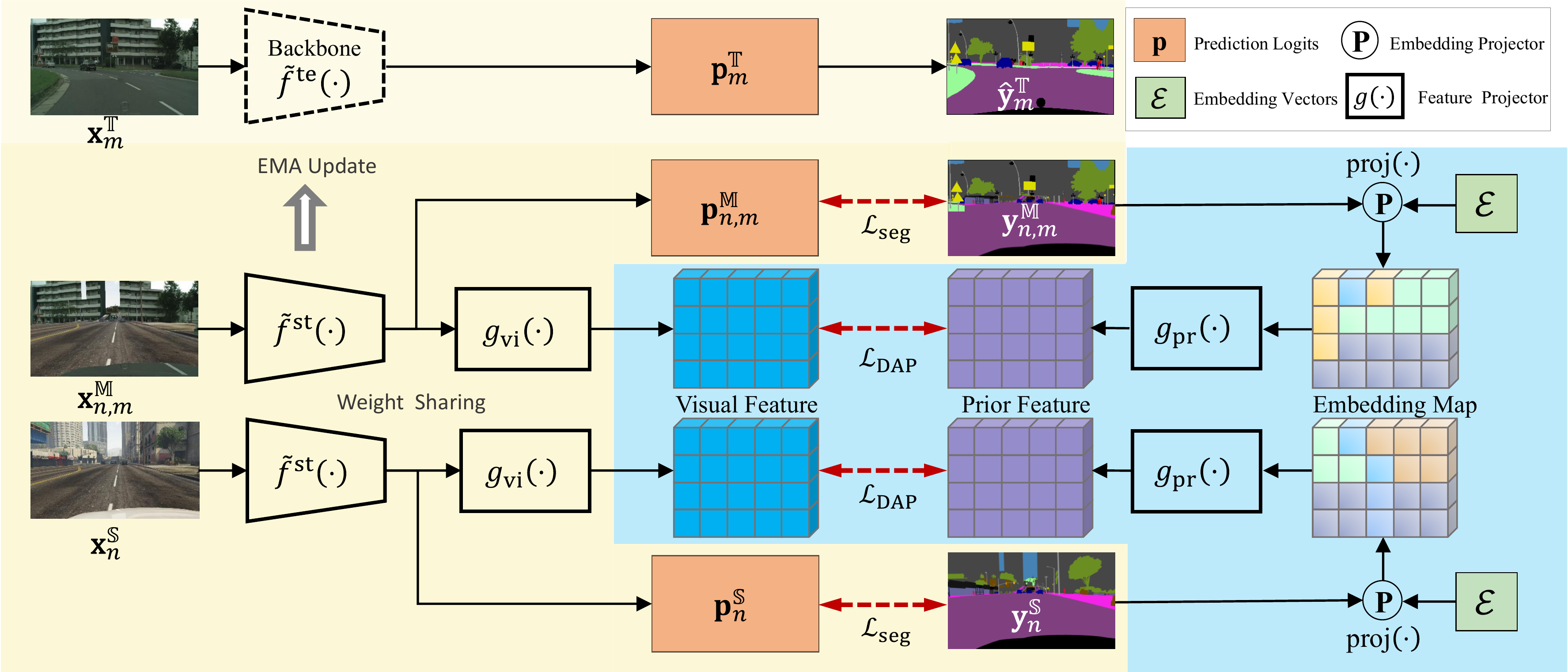}
\caption{The proposed framework of that involves building DAP (the blue-shaded region) upon DACS~\cite{tranheden2021dacs} (the yellow-shaded region). We omit the illustration of using ClassMix~\cite{olsson2021classmix} to generate $\mathbf{x}_{n,m}^\mathbb{T}$. The details of producing the embedding map (\textit{i.e.}, the $\mathrm{proj}(\cdot)$ function) is shown in Fig~\ref{fig:embedding}. \textit{This figure is best viewed in color.}}
\label{fig:framework}
\end{figure*}

\section{Related Work}
\label{related}

Unsupervised domain adaptation (UDA) aims to transfer models trained in a known (labeled) source domain to an unknown (unlabeled) target domain~\cite{ganin2015unsupervised,long2015learning,zou2018unsupervised}. It differs from semi-supervised learning~\cite{zhu2005semi,lee2013pseudo} mainly in the potentially significant difference between the labeled and unlabeled data. The past years have witnessed a fast development of UDA , extending the studied task from image classification~\cite{saito2018maximum,chen2019progressive} to fine-scaled visual recognition including detection~\cite{chen2018domain,kim2019self}, segmentation~\cite{tsai2018learning,chao2021rethinking}, person re-identification~\cite{fu2019self,song2020unsupervised}, \textit{etc}. Specifically, segmentation is a good testbed for UDA for at least two reasons. First, compared to classification, pixel-level segmentation requires more sophisticated manipulation of domain transfer. Second, UDA can reduce the annotation cost for segmentation which is often expensive. The recent approaches of UDA segmentation is roughly categorized into three parts,  adversarial learning, self-training, and data generation.

\noindent$\bullet$\quad\textbf{Adversarial learning} is an important tool for domain adaptation~\cite{tzeng2017adversarial,bousmalis2017unsupervised}. When it was applied to UDA segmentation~\cite{tsai2018learning,toldo2020unsupervised}, the segmentation modules are regarded as the generator that produce prediction, and an auxiliary discriminator is trained to judge which domain the inputs are from. By confusing the discriminator, the purpose of domain adaptation is achieved. Extensions beyond this idea include~\cite{kim2020learning} that facilitated the domain alignment from the input image and transferred the low-level representation (\textit{e.g.} the texture and brightness) from the target domain to the source domain, \cite{yang2020fda} that applied a Fourier transform to narrow the low-frequency gap between the two domains, and others.


\noindent$\bullet$\quad The idea of \textbf{self-training} was borrowed from semi-supervised learning~\cite{tarvainen2017mean,rosenberg2005semi,huo2021atso} and few-shot learning~\cite{li2019learning,liu2020prototype}, and was applied well to domain adaptation~\cite{shin2020two}. The key to self-training is to produce high-quality pseudo labels~\cite{lee2013pseudo}, but prediction errors are inevitable especially when the domain gap is significant. Many efforts have been made to alleviate the inaccuracy, including entropy minimization~\cite{zou2019confidence,vu2019advent,chen2019domain} that tried to increase the confidence on unlabeled data, class-balance regulation~\cite{zou2018unsupervised} that improved the prediction probability for hard categories and thus alleviated the burden that minor categories may be suppressed, and~\cite{guo2021metacorrection,zhou2020uncertainty} that reduced the uncertainty of prediction
 to rectify pseudo labels.

\noindent$\bullet$\quad The \textbf{data generation} branch tried to integrate the advantages of adversarial learning and self-training. The idea is to bridge the gap between the source and target domains using a generated domain in which source and target data are mixed. The images can be generated either by performing GAN~\cite{zhu2017unpaired,hoffman2018cycada} or pixel-level mixing~\cite{olsson2021classmix,zhang2017mixup}. Our approach follows the second path which is verified powerful in UDA segmentation~\cite{tranheden2021dacs,gao2021dsp,zhou2021context}, and we introduce a domain-agnostic prior to constrain representation learning on the target domain.

There have been studies of combining visual data with other types of information, in particular, with text data~\cite{kamath2021mdetr,mottaghi2014role,larochelle2008zero}. Powered by pre-training on a large amount of image and text data, either paired~\cite{radford2021learning} or unpaired~\cite{jia2021scaling}, deep neural networks gain the ability of aligning visual and linguistic features in a shared space, hence facilitating cross-modal understanding~\cite{tan2019lxmert,li2020oscar} and retrieval~\cite{wang2017adversarial,feng2014cross}. Recently, the rapid development of Transformers~\cite{vaswani2017attention,dosovitskiy2020image} brings the possibility of unifying image and text data within one framework. Transferring the learned knowledge from one modality to another may cause the setting somewhat similar to zero-shot learning~\cite{bucher2019zero,zhao2017open}. There are also some discussions on safe and/or efficient cross-modal learning, including~\cite{hu2020creating} that introduced knowledge distillation into cross-modal retrieval, and~\cite{baek2021exploiting} that introduced boundary-aware regression and semantic consistency constraint and improved the discrimination for the unlabeled classes.


\section{Our Approach}
\label{approach}

\subsection{Problem Setting and Baselines}
\label{approach:baseline}

Unsupervised domain adaptation (UDA) starts with defining two domains corresponding to different data distributions. In our setting for semantic segmentation, complete pixel-level annotations are available for the source domain, $\mathbb{S}$, but unavailable for the target domain, $\mathbb{T}$, yet we hold an assumption that the class sets on $\mathbb{S}$ and $\mathbb{T}$ are identical. From the data perspective, let the training samples on $\mathbb{S}$ from a set of $\mathcal{D}^\mathbb{S}=\{(\mathbf{x}_n^\mathbb{S},\mathbf{y}_n^\mathbb{S})\}_{n=1}^N$ where both $\mathbf{x}_n^\mathbb{S}$ is a high-resolution image and $\mathbf{y}_n^\mathbb{S}$ of the same size offers pixel-wise semantic labels. Similarly, we have a training set on $\mathbb{T}$ being $\mathcal{D}^\mathbb{T}=\{(\mathbf{x}_m^\mathbb{T})\}_{m=1}^M$ where ground-truth labels are not provided. The goal is to train a model $\mathbf{y}=f(\mathbf{x};\boldsymbol{\theta})$ on both $\mathcal{D}^\mathbb{S}$ and $\mathcal{D}^\mathbb{T}$ so that it works well on a hidden test set $\mathcal{D}^{\mathbb{T}\prime}$ which is also sampled from $\mathbb{T}$.

The major difficulty of UDA comes from the domain gap between $\mathbb{S}$ and $\mathbb{T}$, \textit{e.g.}, in our testing environment, $\mathbb{S}$ corresponds to synthesized data~\cite{richter2016playing,ros2016SYNTHIA} but $\mathbb{T}$ corresponds to real-world data~\cite{cordts2016cityscapes}. The potential differences in lighting, object styles, \textit{etc.}, can downgrade prediction confidence as well as quality on the target domain. One of the popular solutions to bridge the domain gap is to use the source model to generate pseudo labels $\hat{\mathbf{y}}_m^\mathbb{T}$ for $\mathbf{x}_m^\mathbb{T}$ by a mean-teacher model~\cite{tarvainen2017mean}, and uses both $\mathcal{D}^\mathbb{S}$ and $\hat{\mathcal{D}}^\mathbb{T}=\{(\mathbf{x}_m^\mathbb{T},\hat{\mathbf{y}}_m^\mathbb{T})\}_{m=1}^M$, the extended target set, for updating the online student model. We denote the teacher and student models as $f^\mathrm{te}(\mathbf{x};\boldsymbol{\theta}^\mathrm{te})$ and $f^\mathrm{st}(\mathbf{x};\boldsymbol{\theta}^\mathrm{st})$, respectively, where in the mean-teacher algorithm, $f^\mathrm{te}(\cdot)$ is the moving average of $f^\mathrm{st}(\cdot)$.

The above mechanism, though simple and elegant, suffers the unreliability of $\hat{\mathbf{y}}_m^\mathbb{T}$. To alleviate this issue, in a recently published work named DACS~\cite{tranheden2021dacs}, researchers proposed to replace the training data from the target domain with that from a mixed domain, $\mathbb{M}$, where the samples are generated by mixing images at the pixel level guided by class labels~\cite{olsson2021classmix}. In each training iteration, a pair of source and target images with (true or pseudo) labels are sampled and cropped into the same resolution, denoted as $(\mathbf{x}_n^\mathbb{S},\mathbf{y}_n^\mathbb{S},\mathbf{x}_m^\mathbb{T},\hat{\mathbf{y}}_m^\mathbb{T})$. Next, a subset of classes is randomly chosen from $\mathbf{y}_n^\mathbb{S}$ and a binary mask $\mathbf{M}_{n,m}$ of the same size as $\mathbf{x}_n^\mathbb{S}$ is made, with all pixels that $\mathbf{y}_n^\mathbb{S}$ belongs to the subset being $1$ and otherwise $0$. Upon $\mathbf{M}_{n,m}$, a mixed image with its label is defined:
\begin{equation}
\label{eqn:image-mix}
\left\{\begin{array}{l}
\mathbf{x}_{n,m}^\mathbb{M}=\mathbf{x}_n^\mathbb{S}\odot\mathbf{M}_{n,m}+\mathbf{x}_m^\mathbb{T}\odot(\mathbf{1}-\mathbf{M}_{n,m})\\
\mathbf{y}_{n,m}^\mathbb{M}=\mathbf{y}_n^\mathbb{S}\odot\mathbf{M}_{n,m}+\hat{\mathbf{y}}_m^\mathbb{T}\odot(\mathbf{1}-\mathbf{M}_{n,m})
\end{array}
\right.,
\end{equation}
where $\odot$ denotes element-wise multiplication. The student model, $f^\mathrm{st}(\mathbf{x};\boldsymbol{\theta}^\mathrm{st})$, is trained with $(\mathbf{x}_n^\mathbb{S},\mathbf{y}_n^\mathbb{S})$ and $(\mathbf{x}_n^\mathbb{M},\mathbf{y}_n^\mathbb{M})$:
\begin{equation}
\label{eqn:seg-loss}
\mathcal{L}_\mathrm{seg}=\mathcal{L}_\mathrm{CE}(f^\mathrm{st}(\mathbf{x}_n^\mathbb{S};\boldsymbol{\theta}^\mathrm{st}),\mathbf{y}_n^\mathbb{S})+\mathcal{L}_\mathrm{CE}(f^\mathrm{st}(\mathbf{x}_{n,m}^\mathbb{M};\boldsymbol{\theta}^\mathrm{st}),\mathbf{y}_{n,m}^\mathbb{M}),
\end{equation}
where $\mathcal{L}_\mathrm{CE}(\cdot,\cdot)$ is the pixel-wise cross-entropy loss. The teacher model, $f^\mathrm{te}(\mathbf{x};\boldsymbol{\theta}^\mathrm{te})$, is updated with the student model using the exponential moving average (EMA) mechanism, namely, $\boldsymbol{\theta}^\mathrm{te}\leftarrow\boldsymbol{\theta}^\mathrm{te}\cdot\lambda+\boldsymbol{\theta}^\mathrm{st}\cdot(1-\lambda)$, where $\lambda$ controls the window of EMA and is often close to $1.0$. The entire flowchart of DACS is illustrated in the yellow-shaded part of Fig~\ref{fig:framework}.

\subsection{Confusion of Semantically Similar Classes}
\label{approach:Middle Domain}

\begin{table}[!t]
\centering
\resizebox{0.48\textwidth}{!}{%
\setlength{\tabcolsep}{0.12cm}
\begin{tabular}{cc|cc|cc}
\toprule[1pt]
                             &      & \multicolumn{2}{c|}{$\mathrm{A}$: \textit{motor}, $\mathrm{B}$: \textit{bike}}                    & \multicolumn{2}{c}{$\mathrm{A}$: \textit{road}, $\mathrm{B}$: \textit{sidewalk}}                 \\ \hline
\multicolumn{1}{c|}{Methods} & Data & \multicolumn{1}{c|}{$\mathrm{cos}\langle\boldsymbol{\mu}_\mathrm{A}^{\cdot},\boldsymbol{\mu}_\mathrm{B}^{\cdot}\rangle$} & IOU (\%)        & \multicolumn{1}{c|}{$\mathrm{cos}\langle\boldsymbol{\mu}_\mathrm{A}^{\cdot},\boldsymbol{\mu}_\mathrm{B}^{\cdot}\rangle$} & IOU (\%)       \\ \hline \hline
\multicolumn{1}{c|}{\multirow{2}{*}{\begin{tabular}[c]{@{}c@{}}Source\\ Only\end{tabular}}} & $\mathbb{S}$ & \multicolumn{1}{c|}{\textit{0.42}} & \textit{6.36} & \multicolumn{1}{c|}{\textit{0.35}} & \textit{1.43} \\ \cline{2-6} 
\multicolumn{1}{c|}{}        & $\mathbb{T}$    & \multicolumn{1}{c|}{0.60}             & 27.76          & \multicolumn{1}{c|}{0.43}             & 5.53          \\ \hline
\multicolumn{1}{c|}{DACS}    & $\mathbb{T}$    & \multicolumn{1}{c|}{0.60}             & 26.62          & \multicolumn{1}{c|}{0.38}             & 1.95          \\ \hline
\multicolumn{1}{c|}{DAP}     & $\mathbb{T}$    & \multicolumn{1}{c|}{\textbf{0.56}}    & \textbf{22.19} & \multicolumn{1}{c|}{\textbf{0.23}}    & \textbf{1.29} \\ 
\hline
\end{tabular}}
\vspace{0cm}
\caption{Statistics of features extracted two pairs of semantically similar classes, where $\mathbb{S}$ and $\mathbb{T}$ corresponds to GTAv and Cityscapes, respectively. Please refer to the main text for details.}
\label{tab:feature_sta}
\end{table}

Despite the effectiveness in stabilizing self-learning, DACS still has difficulties in distinguishing semantically similar classes, especially when these classes do not appear frequently in the target domain, \textit{e.g.},  \textit{motorbike} accounts for only $0.1\%$ of the total number of pixels. Fig~\ref{fig:motivation} shows an example that the class pair of \textit{motorbike} and \textit{bike} can easily confuse the model, and so can the pair of \textit{road} and \textit{sidewalk}. According to experimental results, such confusion contributes significantly to segmentation error, \textit{e.g.}, $20.8\%$ mis-classification of \textit{bike} pixels goes to \textit{motorbike}.

We offer a hypothesis for the above phenomenon. Since data from the target domain, say $\mathbf{x}_m^\mathbb{T}$, are not labeled, the semantic correspondence is learned by mapping $\mathbf{x}_m^\mathbb{T}$ to the source domain, \textit{e.g.}, by image-level style transfer for GAN-based approaches~\cite{zhu2017unpaired,hoffman2018cycada} and by label-level simulation for DACS. Mathematically, this is to learn a transfer function (which transfers visual features from the target domain to the source domain) in a weakly-supervised manner. This causes approximation of visual representation and consequently incurs inaccuracy of recognition.

In addition, we consider two semantically similar classes denoted as $\mathrm{A}$ and $\mathrm{B}$, and the corresponding features extracted from these classes form two sets of $\mathcal{F}_\mathrm{A}^\cdot$ and $\mathcal{F}_\mathrm{B}^\cdot$, respectively, where the superscript can be either $\mathbb{S}$ or $\mathbb{T}$. Let us assume that each feature set follows a multi-variate Gaussian distribution denoted by $\mathcal{N}(\boldsymbol{\mu}_\mathrm{A}^\cdot,\boldsymbol{\Sigma}_\mathrm{A}^\cdot)$ and $\mathcal{N}(\boldsymbol{\mu}_\mathrm{B}^\cdot,\boldsymbol{\Sigma}_\mathrm{B}^\cdot)$, abbreviated as $\mathcal{N}_\mathrm{A}^\cdot$ and $\mathcal{N}_\mathrm{B}^\cdot$, respectively. Since $\mathrm{A}$ and $\mathrm{B}$ are semantically similar, we assume that $\boldsymbol{\mu}_\mathrm{A}^\mathbb{S}$ and $\boldsymbol{\mu}_\mathrm{B}^\mathbb{S}$ are close in the feature space, and the source model learns to distinguish $\mathrm{A}$ from $\mathrm{B}$ by reducing $\cos\langle\boldsymbol{\mu}_\mathrm{A}^\mathbb{S},\boldsymbol{\mu}_\mathrm{B}^\mathbb{S}\rangle$ and hence the IOU between $\mathcal{N}_\mathrm{A}^\mathbb{S}$ and $\mathcal{N}_\mathrm{B}^\mathbb{S}$ \footnote{To calculate the IOU between $\mathcal{N}_\mathrm{A}^\mathbb{S}$ and $\mathcal{N}_\mathrm{B}^\mathbb{S}$, we sample an equal amount of points from $\mathcal{N}_\mathrm{A}^\mathbb{S}$ and $\mathcal{N}_\mathrm{B}^\mathbb{S}$, and calculate $r_\mathrm{A}$ as the probability that a point sampled from $\mathcal{N}_\mathrm{A}^\mathbb{S}$ actually has a higher density at $\mathcal{N}_\mathrm{B}^\mathbb{S}$, and $r_\mathrm{B}$ vice versa. The IOU is then approximated as $(r_\mathrm{A}+r_\mathrm{B})/(2-r_\mathrm{A}-r_\mathrm{B})$.}. However, in the target domain, these conditions are not necessarily satisfied since no strong supervision is present -- the distance between $\boldsymbol{\mu}_\mathrm{A}^\mathbb{T}$ and $\boldsymbol{\mu}_\mathrm{B}^\mathbb{T}$ gets smaller (\textit{i.e.}, $\cos\langle\boldsymbol{\mu}_\mathrm{A}^\mathbb{T},\boldsymbol{\mu}_\mathrm{B}^\mathbb{T}\rangle$ gets larger), and, consequently, the IOU between $\mathcal{N}_\mathrm{A}^\mathbb{T}$ and $\mathcal{N}_\mathrm{B}^\mathbb{T}$ is larger. Tab~\ref{tab:feature_sta} offers quantitative results from transferring a segmentation model from GTAv to Cityscapes. The features of two semantically similar class pairs, namely, \textit{motorbike} vs. \textit{bike}, and \textit{road} vs. \textit{sidewalk}, are both made easier to confuse the network.

\subsection{Domain-Agnostic Prior for UDA Segmentation}
\label{approach:prior}

To obtain more accurate estimation for $\mathcal{N}_\mathrm{A}^\mathbb{T}$ and $\mathcal{N}_\mathrm{B}^\mathbb{T}$, we refer to the Bayesian theory that the posterior distribution is composed of a prior and a likelihood. In our setting, the likelihood comes from the target dataset, $\mathcal{D}^\mathbb{T}$, where there are insufficient data to guarantee accurate estimation. The solution lies in introducing an informative prior for the elements, namely, $\mathbf{z}_\mathrm{A}^\mathbb{T}\sim\mathcal{N}_\mathrm{A}^\mathbb{T}$ and $\mathbf{z}_\mathrm{B}^\mathbb{T}\sim\mathcal{N}_\mathrm{B}^\mathbb{T}$. This involves defining constraints as follows:
\begin{equation}
\label{eqn:prior}
\mathbf{z}_\mathrm{A}^\mathbb{T}=g(\mathbf{e}_\mathrm{A}),\quad\mathbf{z}_\mathrm{B}^\mathbb{T}=g(\mathbf{e}_\mathrm{B}),
\end{equation}
where $\mathbf{e}_\mathrm{A}$ and $\mathbf{e}_\mathrm{B}$ are \textbf{not} related to any specific domain -- we name them \textbf{domain-agnostic priors} (DAP), and $g(\cdot)$ is a learnable function that projects the priors to the semantic space. In practice, it is less likely to strictly satisfy Eqn~\eqref{eqn:prior}, so we implement it as a loss term to minimize.

We instantiate DAP into two examples. The first one is a set of one-hot vectors, \textit{i.e.}, for a target domain with $C$ classes, the $c$-th class is encoded into $\mathbf{I}_c$, a $C$-dimensional vector in which all entries are $0$ except for the $c$-th dimension being $1$. The second one is borrowed from word2vec~\cite{mikolov2013distributed}, a pre-trained language model that represents each word using a $300$-dimensional vector. Note that both cases are completely independent from the vision domain, \textit{i.e.}, reflecting the principle of being domain-agnostic. We denote the set of embedding vectors as $\mathcal{E}=\{\mathbf{e}_c\}_{c=1}^C$, where $\mathbf{e}_c$ is the embedding vector of the $c$-th class.

Back to the main story, DAP and the embedded vectors are easily integrated into DACS, the baseline framework. Besides the segmentation loss $\mathcal{L}_\mathrm{seg}$ defined in Eqn~\eqref{eqn:seg-loss}, we introduce another loss term $\mathcal{L}_\mathrm{DAP}$ that measures the distance between the embedded domain-agnostic priors and visual features extracted from training images,
\begin{equation}
\begin{aligned}
\mathcal{L}_\mathrm{DAP}=&\|g_\mathrm{vi}(\tilde{f}^\mathrm{st}(\mathbf{x}_n^\mathbb{S};\boldsymbol{\theta}^\mathrm{st}))-g_\mathrm{pr}(\mathrm{proj}(\tilde{\mathbf{y}}_n^\mathbb{S};\mathcal{E}))\|_2^2+\\
&\|g_\mathrm{vi}(\tilde{f}^\mathrm{st}(\mathbf{x}_{n,m}^\mathbb{M};\boldsymbol{\theta}^\mathrm{st}))-g_\mathrm{pr}(\mathrm{proj}(\tilde{\mathbf{y}}_{n,m}^\mathbb{M};\mathcal{E}))\|_2^2,
\end{aligned}
\end{equation}
where $\tilde{f}^\mathrm{st}(\cdot)$ is the backbone part of $f^\mathrm{st}(\cdot)$ that extracts mid-resolution features (\textit{i.e.}, visual features), $\tilde{\mathbf{y}}_n^\mathbb{S}$ and $\tilde{\mathbf{y}}_{n,m}^\mathbb{M}$ denote the supervisions adjusted to the same size of the backbone outputs, $\mathrm{proj}(\cdot)$ projects the domain-agnostic embedding vectors to the image plane based on the labels (to be detailed later), and $g_\mathrm{vi}(\cdot)$ and $g_\mathrm{pr}(\cdot)$ are each a learnable convolution layer that maps initial features from the vision and prior spaces to the common space.

\begin{figure}[!t]
\includegraphics[width=0.45\textwidth]{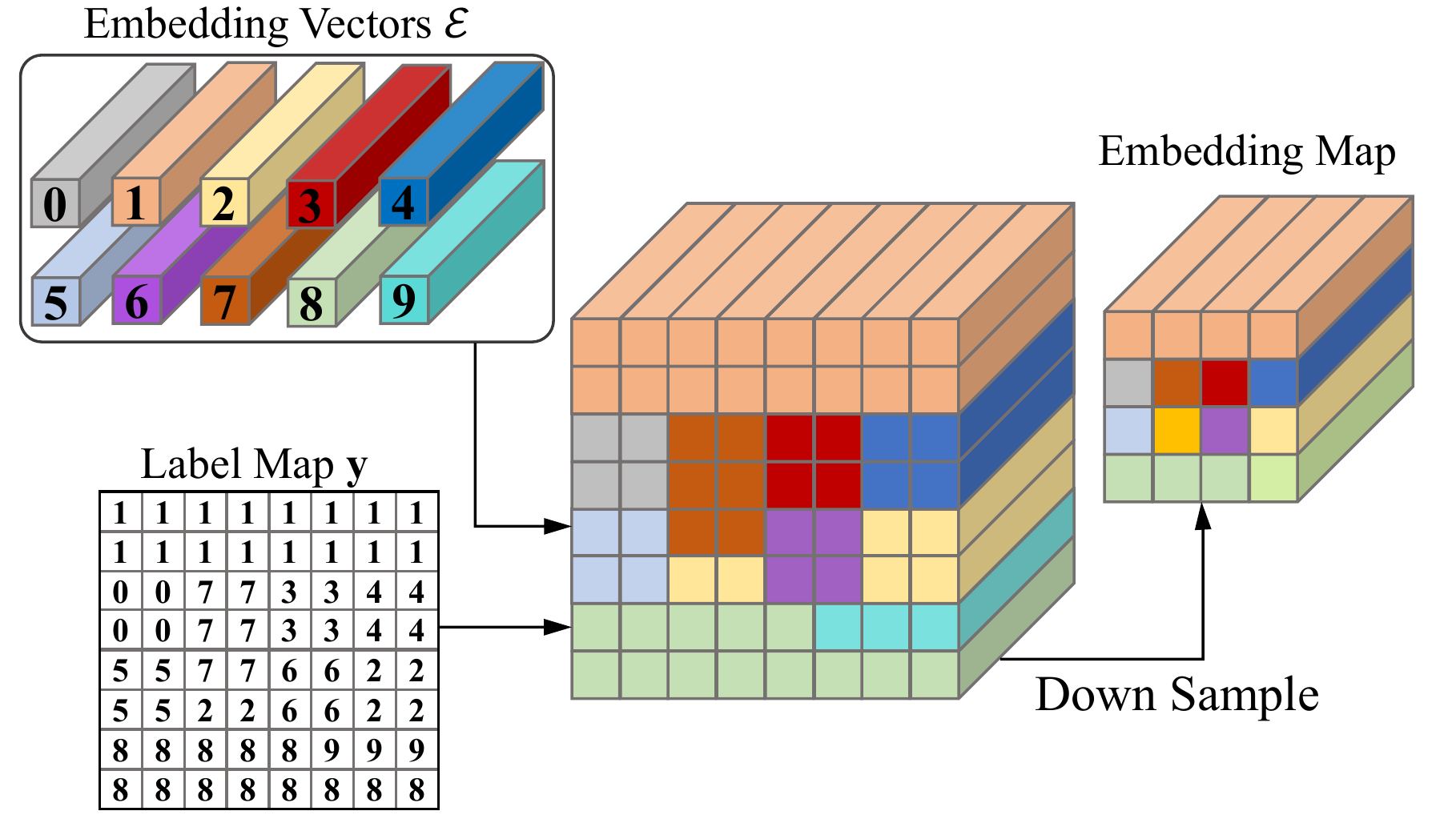}
\caption{An illustration of constructing the domain-agnostic embedding map with prior vectors. Each number in the label map stands for a class ID. The total number of classes can be arbitrary.}
\label{fig:embedding}
\end{figure}

We elaborate the construction of $\mathrm{proj}(\cdot)$ in Fig~\ref{fig:embedding}. The computation is simple: for each position on the image plane, we refer to the ground-truth label $\tilde{\mathbf{y}}_{n}^\mathbb{S}$ or partly-pseudo label $\tilde{\mathbf{y}}_{n,m}^\mathbb{M}$, and directly paste the class-wise embedding vectors to the corresponding positions. Since $\mathcal{E}$ is fixed, this procedure does not need any gradient back-propagation. We are aware that advanced versions of $\mathrm{proj}(\cdot)$ (\textit{e.g.}, performing local smoothing) are available, and we leave these properties to be learned by $g_\mathrm{vi}(\cdot)$ and $g_\mathrm{pr}(\cdot)$.

Finally, the overall loss function is written as
\begin{equation}
\mathcal{L}_\mathrm{overall}=\mathcal{L}_\mathrm{seg}+\alpha\cdot\mathcal{L}_\mathrm{DAP},
\end{equation}
where $\alpha$ is the balancing coefficient which is by default set to be $1.0$ -- an analysis of the effect of $\alpha$ is provided in the experimental part (see Tab~\ref{tab:alpha}).

Back to Tab~\ref{tab:feature_sta}, one can observe how DAP reduces the IOU between $\mathcal{N}_\mathrm{A}^\mathbb{T}$ and $\mathcal{N}_\mathrm{B}^\mathbb{T}$, and thus alleviate the confusion between the semantically similar classes. Interestingly, the gains of two class pairs are consistent and significant, even though the training samples of \textit{road} vs. \textit{sidewalk} being more abundant. In experiments, we shall see how the ability of discriminating these class pairs is improved.

\subsection{Discussions}
\label{approach:discussions}


To the best of our knowledge, this is the first work that integrates text embedding into UDA segmentation and producing considerable accuracy gain, which demonstrating the effectiveness of linguistic cues assisting visual recognition. However, it is yet a preliminary solution, and  some possible directions can be discovered.

\vspace{0.1cm}
\noindent$\bullet$\quad\textbf{Enhancing text embedding.} The currently used word2vec features it does not consider different words that correspond to the same semantics (\textit{e.g.}, \textit{person} can be \textit{pedestrian}). Interestingly, we tried to enhance the prior by searching for semantically similar words, but obtained little accuracy gain. This may call for a complicated mechanism of exploring the text world.

\vspace{0.1cm}
\noindent$\bullet$\quad\textbf{Constructing domain-agnostic yet vision-aware priors.} This is to answer the question: what kind of image data is considered to offer domain-free information? The answer may lie in generalized datasets like ImageNet~\cite{deng2009imagenet} or Conceptual Captioning~\cite{sharma2018conceptual}, or even the pre-trained image-text models such as CLIP~\cite{radford2021learning} that absorbed $400$ million image-text pairs. Note that it is a major challenge to disentangle domain-related information to avoid over-fitting, and we will continue exploring the possibility in the future.

\section{Experiments}
\label{experiments}

\begin{table*}[!t]
\centering
\resizebox{\textwidth}{!}{%
\begin{tabular}{lcccccccccccccccccccc}
\toprule[2pt]
\multicolumn{21}{c}{\textbf{GTAv$\rightarrow$Cityscapes}}                                                               \\ \midrule[1pt]
\multicolumn{1}{c|}{\textbf{Method}} &
  \rotatebox{90}{Road}&
  \rotatebox{90}{Sidewalk} &
  \rotatebox{90}{Building} &
  \rotatebox{90}{Wall} &
 \rotatebox{90}{Fence} &
 \rotatebox{90}{Pole} &
  \rotatebox{90}{Light} &
  \rotatebox{90}{Sign} &
  \rotatebox{90}{Veg} &
  \rotatebox{90}{Terrain} &
 \rotatebox{90}{ Sky }&
  \rotatebox{90}{Person }&
  \rotatebox{90}{Rider} &
  \rotatebox{90}{Car }&
  \rotatebox{90}{Truck} &
  \rotatebox{90}{Bus} &
  \rotatebox{90}{Train} &
  \rotatebox{90}{Motor} &
  \multicolumn{1}{c|}{\rotatebox{90}{Bike}} &
  \textbf{mIOU} \\ \hline\hline
  
\multicolumn{1}{c|}{AdaptSegNe~\cite{tsai2018learning}}            & 86.5 & 36.0 & 79.9 & 23.4 & 23.3 & 23.9 & 35.2 & 14.8 & 83.4 & 33.3 & 75.6 & 58.5 & 27.6 & 73.7 & 32.5 & 35.4 & 3.9 & 30.1 &\multicolumn{1}{c|} {28.1} & 42.4 \\  

\multicolumn{1}{c|}{PatchAligne~\cite{tsai2019domain}}            & 92.3 & 51.9 &82.1 & 29.2 & 25.1 & 24.5 & 33.8 & 33.0 & 82.4 & 32.8 &82.2 & 58.6 & 27.2 & 84.3 & 33.4& 46.3 & 2.2 & 29.5 &\multicolumn{1}{c|} {32.3} & 46.5 \\

\multicolumn{1}{c|}{LTIR~\cite{kim2020learning}}            &  92.9 & 55.0 & 85.3 & 34.2 & 31.1 & 34.9 & 40.7 & 34.0 & 85.2 & 40.1 & 87.1 & 61.0 & 31.1 & 82.5 & 32.3 & 42.9 & 0.3 & 36.4 & \multicolumn{1}{c|}{ 46.1} & 50.2\\

\multicolumn{1}{c|}         {PIT~\cite{lv2020cross}}  & 87.5 & 43.4 & 78.8 & 31.2 & 30.2 & 36.3 & 39.9 & 42.0 & 79.2 & 37.1 & 79.3 & \underline{65.4} & \textbf{37.5} & 83.2 & \underline{46.0} & 45.6 & \textbf{25.7} & 23.5 &  \multicolumn{1}{c|} {\underline{49.9} }& 50.6   \\

\multicolumn{1}{c|}{FDA~\cite{yang2020fda}}            & 92.5 & 53.3 & 82.4 & 26.5 & 27.6 & 36.4 & 40.6 & 38.9 & 82.3 & 39.8 & 78.0 & 62.6 & 34.4 & 84.9 & 34.1 & \textbf{53.1} & 16.9 & 27.7 &\multicolumn{1}{c|} {46.4} & 50.5 \\

\multicolumn{1}{c|}{MetaCorrect~\cite{guo2021metacorrection} }           & 92.8 & \underline{58.1} & 86.2 & \underline{39.7} & 33.1 & 36.3 & 42.0 & 38.6 & 85.5 & 37.8 & 87.6 & 62.8 & 31.7 & 84.8 & 35.7 & 50.3 & 2.0 & 36.8 &\multicolumn{1}{c|} {48.0} & 52.1   \\

\multicolumn{1}{c|}{DACS~\cite{tranheden2021dacs}} & 89.9 & 39.7 & \textbf{87.9} &  \underline{39.7}& \textbf{39.5} & 38.5 & \underline{46.4} & \textbf{52.8} & \textbf{88.0} & \underline{44.0} & \textbf{88.8} & 67.2 & 35.8 & 84.5 & 45.7 & 50.2 & 0.0 & 27.3 & \multicolumn{1}{c|}{34.0} & 52.1  \\ 

\multicolumn{1}{c|}{IAST~\cite{mei2020instance}}            &  \textbf{94.1} & \textbf{58.8} & 85.4& \underline{39.7}& 29.2 & 25.1 & 43.1 & 34.2 & 84.8 & 34.6 & \underline{88.7} & 62.7 & 30.3 & 87.6 & 42.3 & 50.3 & 24.7 &35.2 &\multicolumn{1}{c|}{40.2} & 52.2 \\

\multicolumn{1}{c|}{DPL~\cite{cheng2021dual}}    & 92.8 & 54.4 & 86.2 & \textbf{41.6} & 32.7 & 36.4 & \textbf{49.0} & 34.0 & 85.8 & 41.3 & 86.0 & 63.2 & 34.2 & 87.2 & 39.3 & 44.5 &\underline{18.7} & \textbf{42.6} &  \multicolumn{1}{c|}{43.1} & 53.3 \\ \hline

\multicolumn{1}{c|}{Source only} &  75.6&  17.1&69.8  & 10.7 & 16.1 &21.1  &27.0 &10.6 &77.3  &15.1 &71.1 & 53.4  &20.5  &73.9 &28.6   & 31.1 & 1.62 & 32.4 & \multicolumn{1}{c|}{ 21.5}&  35.5\\
\multicolumn{1}{c|}{DACS (rep)} & 93.1 & 48.1 & 87.3 &36.7 &35.1  &\underline{38.7} &42.5  &49.3  &\underline{87.5} &41.9 & 87.9 &64.8  &30.7  &\underline{88.3}  &40.2  &51.0  &0.0  & 25.1   &\multicolumn{1}{c|}{42.6} & 52.1 \\
\multicolumn{1}{c|}{DACS+DAP} & \underline{93.5} & 53.9 & \underline{87.5} & 30.0 & \underline{36.4} & \textbf{39.0}  & 43.9 &\underline{49.5} & \underline{87.5} & \textbf{45.4} & \textbf{88.8} & \textbf{66.6} & \underline{36.8} & \textbf{89.4} & \textbf{49.1} & \underline{51.4} & 0.0 & \underline{42.2}& \multicolumn{1}{c|}{\textbf{53.1}} & \textbf{55.0} \\ \hline\hline

\multicolumn{1}{c|}{ProDA~\cite{zhang2021prototypical}}    & 87.8 & 56.0 & 79.7 & {46.3} & {44.8} & {45.6} & {53.5} & {53.5} & {88.6} & {45.2} & 82.1 & {70.7} & {39.2} & {88.8} & {45.5} & {50.4} & 1.0 & {48.9} &  \multicolumn{1}{c|}{56.4} & 57.5 \\

\multicolumn{1}{c|}{Chao \textit{et al.}~\cite{chao2021rethinking}}    & 94.4 & 60.9 & 88.1 & 39.5 & 41.8 & 43.2 & 49.1 & 56.0 & 88.0 & 45.8 & 87.8 & 67.6 & 38.1 & 90.1 & 57.6 & 51.9 & 0.0 & 46.6 & \multicolumn{1}{c|}{55.3}& 58.0\\

\multicolumn{1}{c|}{DAP + ProDA}    & \textbf{94.5} & \textbf{63.1} & \textbf{89.1} & 29.8 & \textbf{47.5} & \textbf{50.4} & \textbf{56.7} & \textbf{58.7} & \textbf{89.5} & \textbf{50.2} & 87.0 & \textbf{73.6} & 38.6 & \textbf{91.3} & 50.2 & \textbf{52.9} & 0.0 & \textbf{50.2} &  \multicolumn{1}{c|}{\textbf{63.5}} & \textbf{59.8 }\\

\bottomrule[2pt]
\end{tabular}}
\vspace{0cm}
\caption{Segmentation accuracy (IOU, \%) of different UDA approaches from GTAv to Cityscapes. DACS (rep) indicates our re-implementation of DACS. For each class, we mark the highest number with \textbf{bold} and the second highest with \underline{underline}. The top and bottom parts are achieved without and with  multi-stage training or multi-model fusion.}
\label{GTA}
\end{table*}

\begin{table*}[!t]
\centering
\resizebox{\textwidth}{!}{
\begin{tabular}{lcccccccccccccccccc}
\toprule[2pt]
\multicolumn{19}{c}{\textbf{SYNTHIA$\rightarrow$Cityscapes}}                                                                              \\ \midrule[1pt]
\multicolumn{1}{c|}{Method} &
 \rotatebox{90}{Road}&
  \rotatebox{90}{Sidewalk} &
  \rotatebox{90}{Building} &
  \rotatebox{90}{Wall*} &
 \rotatebox{90}{Fence*} &
 \rotatebox{90}{Pole*} &
  \rotatebox{90}{Light} &
  \rotatebox{90}{Sign} &
  \rotatebox{90}{Veg} &
 \rotatebox{90}{ Sky }&
  \rotatebox{90}{Person }&
  \rotatebox{90}{Rider} &
  \rotatebox{90}{Car }&
  \rotatebox{90}{Bus} &  
  \rotatebox{90}{Motor} &
  \multicolumn{1}{c|}{\rotatebox{90}{Bike} }&
  \multicolumn{1}{l|}{mIOU} &
  mIOU* \\ \hline\hline
\multicolumn{1}{c|}{PatchAlign~\cite{tsai2019domain}}            & 82.4& 38.0 & 78.6 & 8.7 & 0.6 & 26.0 & 3.9 & 11.1 & 75.5 & 84.6 & 53.5 & 21.6 & 71.4 & 32.6 & 19.3 &31.7 & \multicolumn{1}{|c|}{40.0} & 46.5 \\    
  
\multicolumn{1}{c|}{AdaptSegNe~\cite{tsai2018learning}}            & 84.3 & 42.7 & 77.5 & -- & -- & -- & 4.7 & 7.0 & 77.9 & 82.5 & 54.3 & 21.0 & 72.3 & 32.2 & 18.9 & 32.3 & \multicolumn{1}{|c|}{--} & 46.7 \\     
  
\multicolumn{1}{c|}{FDA~\cite{yang2020fda}}            &  79.3 & 35.0 & 73.2 & -- & -- & -- & 19.9 & 24.0 & 61.7 & 82.6 & 61.4 & 31.1 & 83.9 & 40.8 & \textbf{38.4} & \underline{51.1} & \multicolumn{1}{|c|}{--} & 52.5 \\

\multicolumn{1}{c|}{LTIR~\cite{kim2020learning}}            & \textbf{92.6} & \textbf{53.2} & 79.2 & -- & -- & -- & 1.6 & 7.5 & 78.6 & 84.4 & 52.6 & 20.0 & 82.1 & 34.8 & 14.6 & 39.4 & \multicolumn{1}{|c|}{--} & 49.3  \\

\multicolumn{1}{c|}{PIT~\cite{lv2020cross}}            & 83.1 & 27.6 & 81.5 & 8.9 & 0.3 & 21.8 & 26.4 & \textbf{33.8} & 76.4 & 78.8 & 64.2 & 27.6 & 79.6 & 31.2 & 31.0 & 31.3 &  \multicolumn{1}{|c|}{44.0} & 51.8 \\

\multicolumn{1}{c|}{MetaCorrect~\cite{guo2021metacorrection}}            & 92.6 & \underline{52.7} & 81.3 & 8.9 & 2.4 & 28.1 & 13.0 & 7.3 & 83.5 & 85.0 & 60.1 & 19.7 & \underline{84.8} & 37.2 & 21.5 & 43.9 & \multicolumn{1}{|c|}{45.1}  & 52.5\\

\multicolumn{1}{c|}{DPL~\cite{cheng2021dual}}  &87.5 & 45.7 & \underline{82.8} & 13.3 & 0.6 & 33.2 & 22.0 & 20.1 & 83.1 & 86.0 & 56.6 & 21.9 & 83.1 & 40.3 & 29.8&\multicolumn{1}{c|}{45.7} & \multicolumn{1}{|c|}{47.0} &  54.2\\

\multicolumn{1}{c|}{DACS~\cite{tranheden2021dacs}}            & 80.6 & 25.1 & 81.9 & \underline{21.5} &\underline{ 2.9} & \textbf{37.2} & 22.7 & 24.0 & 83.7 & \textbf{90.8} &\textbf{67.6} & \textbf{38.3} & 82.9 & 38.9 & 28.5 & 47.6 & \multicolumn{1}{|c|}{48.3} & 54.8 \\ 

\multicolumn{1}{c|}{IAST~\cite{mei2020instance}}    & 81.9 & 41.5 & \textbf{83.3} & 17.7 & \textbf{4.6} & 32.3 & \underline{30.9} & \underline{28.8} & 83.4 & 85.0 & 65.6 & 30.8 & \textbf{86.5}& 38.2 & 33.1 & \textbf{52.7} & 
\multicolumn{1}{|c|}{49.8} & 57.0 \\\hline

\multicolumn{1}{c|}{Source only} &  24.7&  12.1& 75.4 &11.3  &0.1  &22.4  & 7.5 &16.6  & 71.7 &  78.3 & 52.9 & 10.1 & 56.6 & 23.3 & 4.0 & 13.0 & \multicolumn{1}{|c|}{30.0} & 34.3 \\

\multicolumn{1}{c|}{DACS (rep)}        &  82.1&31.0  &82.4  & 22.1 & 1.2 & 33.1 & \textbf{32.7} &25.1  & \textbf{84.4} & 88.2 & 65.2 & 34.3 & 83.4 & \underline{42.9} &24.1  & 50.8 & \multicolumn{1}{|c|}{48.9} & 55.9 \\
\multicolumn{1}{c|}{DACS+DAP}        & \underline{83.9} & 33.3 & 80.2 &  \textbf{24.1}& 1.2 & \underline{33.4} & 30.8 &\textbf{33.8}  & \underline{84.3} & \underline{88.5} & \underline{65.7} & \underline{36.2} & 84.3 & \textbf{43.3} & \underline{33.3} &  46.3 & \multicolumn{1}{|c|}{\textbf{50.2}} &  \textbf{57.2} \\ \hline\hline

\multicolumn{1}{c|}{Chao \textit{et al.}~\cite{chao2021rethinking}}    & 88.7 & 46.7 & 83.8 &22.7 & 4.1& 35.0 &35.9& 36.1& 82.8 & 81.4 & 61.6 &32.1 & 87.9 & 52.8 & 32.0 & 57.7&
\multicolumn{1}{|c|}{52.6} & 60.0 \\ 

\multicolumn{1}{c|}{ProDA~\cite{zhang2021prototypical}} &   87.8 & 45.7 & 84.6 & {37.1} & 0.6 & 44.0 & {54.6} & {37.0} & {88.1} & 84.4 & 74.2 & 24.3 & 88.2 & {51.1} & {40.5} & 45.6 & 
\multicolumn{1}{|c|}{55.5} & 62.0 \\
\multicolumn{1}{c|}{DAP + ProDA} &  84.2& \textbf{46.5} & 82.5 & 35.1 & 0.2 & \textbf{46.7} & {53.6} & \textbf{45.7} & \textbf{89.3} & \textbf{87.5} & \textbf{75.7} & \textbf{34.6} & \textbf{91.7} & \textbf{73.5} & \textbf{49.4} & \textbf{60.5} & 
\multicolumn{1}{|c|}{\textbf{59.8}} & \textbf{64.3} \\

\bottomrule[2pt]
\end{tabular}}
\vspace{0cm}
\caption{Segmentation accuracy (IOU, \%) of different UDA approaches from SYNTHIA to Cityscapes. DACS (rep) indicates our re-implementation of DACS. For each class, we mark the highest number with \textbf{bold} and the second highest with \underline{underline}. The top and bottom parts are achieved without and with  multi-stage training or multi-model fusion. mIOU* denotes the average of 13 classes, with the classes marked with * not computed.}
\label{SYNTHIA}
\end{table*}

\subsection{Datasets and Implementation Details}
\label{experiments:setting}

\noindent$\bullet$\quad\textbf{Datasets.} We evaluate our method on a popular scenario transferring the information from a synthesis domain to a real domain. We use  GTAv~\cite{richter2016playing} and SYNTHIA~\cite{ros2016SYNTHIA} as composite domain datasets and Cityscapes~\cite{cordts2016cityscapes} as the real domain. GTAv ~\cite{richter2016playing} is a synthetic dataset extracted from the game of Grand Theft Auto V. There are $24\rm{,}966$ images with pixel-level semantic segmentation ground truth. The resolution of these images is $1914\times1052$ and we resize them into $1280 \times 720$ in our experiments. GTAv shares $19$ common classes with Cityscapes. SYNTHIA~\cite{ros2016SYNTHIA} contains $9\rm{,}400$ virtual European-style urban images whose resolution is $1280\times760$ and we keep the original size in our experiments. We 
evaluate two settings (13 and 16 categories) in SYNTHIA. Cityscapes is a large-scale dataset  with a resolution of $2048\times1024$.  There are $2\rm{,}975$ and $500$ images in the training and validation sets, respectively.



\vspace{0.1cm}
\noindent$\bullet$\quad\textbf{Implementation Details.} To be consistent with other methods, we use the Deeplabv2~\cite{chen2017deeplab}
framework with a RseNet101~\cite{he2016deep} backbone as our image encoder and an ASPP classifier as segmentation head. The output map is up-sampled and operated by a softmax layer to match the size of the inputs. The pre-trained model on ImageNet~\cite{deng2009imagenet} and MSCOCO~\cite{lin2014microsoft} is applied to initialize the backbone. The visual and prior feature projectors, $g_\mathrm{vi}(\cdot)$  and  $g_\mathrm{pr}(\cdot)$, are both  convolution layers with a $1 \times 1$ kernel to adjust the channel to $256$.  The batch size of one GPU is set to $2$. We use SGD with Nesterov acceleration as the optimizer which is decreased based on a polynomial decay policy with exponent $0.9$. The initial learning rate of backbone and feature projectors is $2.5 \times 10^{-4}$, that of segmentation head is $10\times$ larger. The momentum and weight decay of the optimizer are $0.9$ and $5\times10^{-4}$. During the training process, we apply color jittering and Gaussian blurring on the mixed data and only resize operation on source and target data.  Teacher model is updated with an EMA decay and $\lambda$ equals $0.99$. We train the model for $250\mathrm{K}$ iterations on a single NVIDIA Tesla-V100 GPU and adopt an early stop setting. The background and invalid categories are ignored during training. The weight of $\mathcal{L}_\mathrm{DAP}$, \textit{i.e.} $\alpha$, is set to be $1.0$. We adapt bilinear interpolation to down-sample the embedding map from the input size to that of the visual feature map -- this is an important step for DAP, which is ablated in Section~\ref{experiments:diagnosis}.

\begin{figure*}[!t]
\centering
\includegraphics[width=1.0\textwidth]{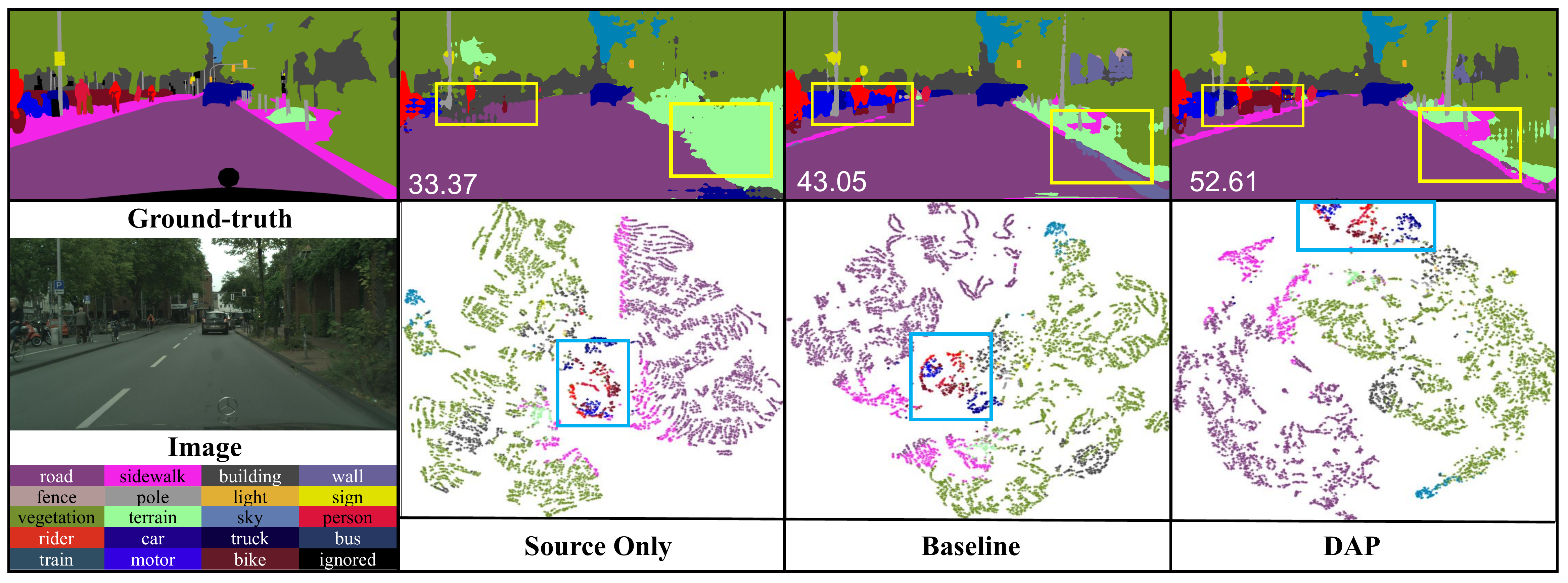}
\caption{An example of transfer segmentation from GTAv to Cityscapes. The top row shows the ground-truth and three segmentation results, while the input and legend is in the bottom row. The yellow boxes indicate the regions that the segmentation quality is largely improved, and the number corresponds to single-image mIOU (over the existing classes). The bottom row also shows the t-SNE of visual features colored by the predicted class. The blue boxes locate the features of \textit{bike} and \textit{motorbike}, in which DAP achieves a favorable ability to distinguish them. \textit{This figure is best viewed in color and we suggest the reader to zoom in for details.}}
\label{fig:comparison}
\end{figure*}

\subsection{Quantitative Results and Visualization}
\label{experiments:quantitative}

We first evaluate DAP on the transfer segmentation task from GTAv to Cityscapes. The comparison against recent approaches are shown in Tab~\ref{GTA}. To ensure reliability, we run DACS (the baseline) and DAP three times and report the average accuracy. DAP achieves a mean IOU of $55.0\%$ over 19 classes, which claims a $2.9\%$ gain beyond the baseline and also outperforms all other competitors except for Chao \textit{et al.}~\cite{chao2021rethinking} and ProDA~\cite{zhang2021prototypical}. Specifically, Chao \textit{et al.}~\cite{chao2021rethinking} used ensemble learning to integrate the prediction from four complementarily-trained models, including DACS, but DAP used a single model; ProDA~\cite{zhang2021prototypical} improved the segmentation accuracy significantly via multi-stage training, yet its first stage reported a $53.7\%$ mIOU.What is more, the results on transferring SYNTHIA to Cityscapes, as shown in Tab~\ref{SYNTHIA}, demonstrate the similar trend -- DAP outperforms all the competitors, except for ProDA and RED, in terms of either 13-class or 16-class mIOU.  To show that DAP offers complementary benefits, we feed the output of DAP as the pseudo labels to the 1st stage of ProDA, and the 2nd and 3rd stages remain unchanged. As shown in Tables~\ref{GTA} and~\ref{SYNTHIA}, the segmentation mIOUs of ProDA in GTAv$\rightarrow$Cityscapes and SYNTHIA$\rightarrow$Cityscapes are improved by $2.3\%$ and $4.3\%$, respectively, setting new records in these two  scenarios.



Next, we investigate the ability of DAP in distinguishing semantically similar classes. From GTAv to Cityscapes, the segmentation mIOUs of \textit{bike} and \textit{motorbike} are improved from $42.6\%$ and $25.1\%$ to $53.1\%$ and $42.2\%$, with absolute gains of $10.5\%$ and $17.1\%$, respectively. From SYNTHIA to Cityscapes, the mIOU of \textit{bike} drops by $4.5\%$ and that of \textit{motorbike} increases by $9.2\%$, achieving an average improvement of $2.4\%$. We visualize an example of segmentation in Fig~\ref{fig:comparison}. Besides a qualitative observation on the improvement of distinguishing \textit{bike} from \textit{motorbike} and \textit{road} from \textit{sidewalk}, we also notice the reason behind the improvement being a scattered feature distribution of these similar classes. This aligns with the statistics shown in Tab~\ref{tab:feature_sta}, indicating that DAP reduces the IOU between the estimated distributions of \textit{bike} and \textit{motorbike} as well as that between \textit{road} from \textit{sidewalk}.

\begin{figure}[!t]
\centering
\includegraphics[width=0.5\textwidth]{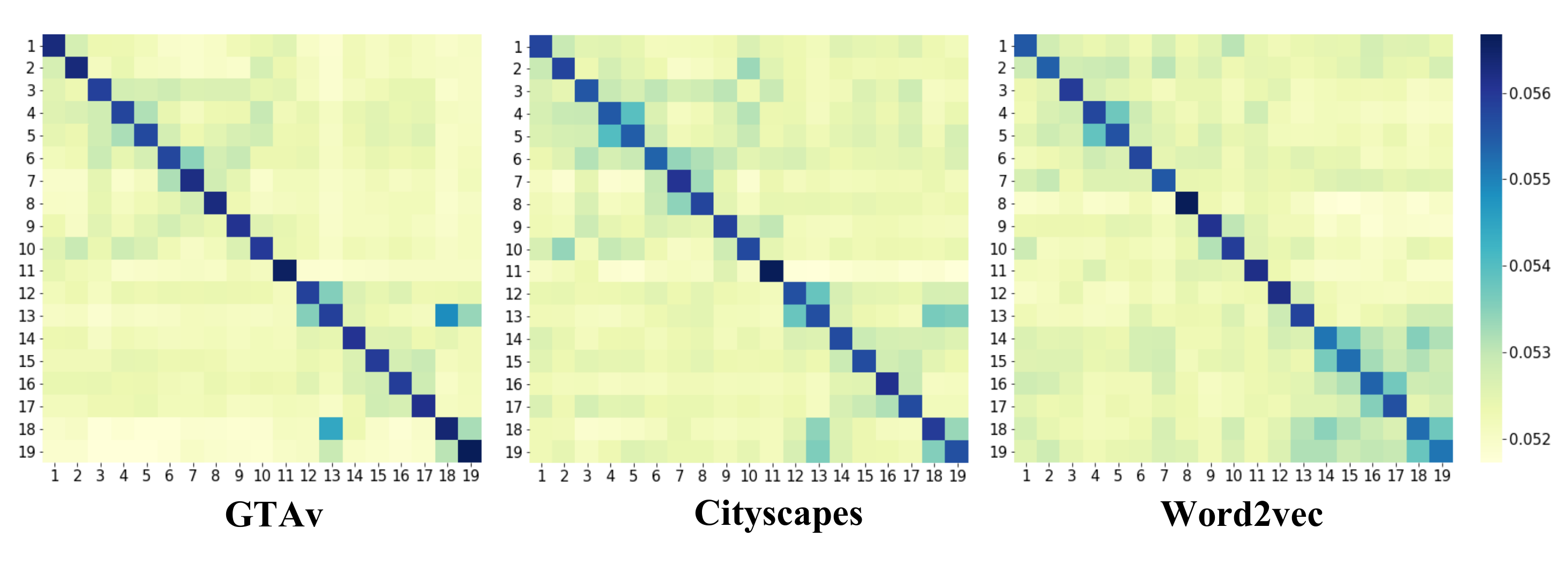}
\caption{The relationship matrix of all 19 classes, where the left and middle are generated by the class-averaged feature vectors extracted from the models directly trained on GTAv and Cityscapes, and the right uses the word2vec features. Each cell is the inner product of two normalized features. The order of class is identical to that in Tab~\ref{GTA} -- for easier reference: 1 for \textit{road}, 2 for \textit{sidewalk}, 12 for \textit{person}, 13 for \textit{rider}, \textit{18} for \textit{motorbike}, 19 for \textit{bike}.}
\label{fig:relationship}
\end{figure}

From another perspective, we study how the language-based prior assists visual recognition. In Fig~\ref{fig:relationship}, we show the relationship matrix of the features learned from the source (GTAv) and target (Cityscapes) domains as well as the word2vec features. There is an interesting example that \textit{person} and \textit{rider} are semantically similar in both GTAv and Cityscapes but not so correlated according to word2vec. This may cause a strong yet harmful correlation of these two feature sets, leading to confusion in segmentation. The relatively weaker correlation of word2vec features alleviates the bias and improves the IOU of both classes (refer to Tab~\ref{GTA}). Similar phenomena are also observed for the class pairs of \textit{road} vs. \textit{sidewalk}, and \textit{bike} vs. \textit{motorbike}.

\begin{table}[!t]
\centering
\resizebox{0.48\textwidth}{!}{
\setlength{\tabcolsep}{0.08cm}
\begin{tabular}{c|ccc|ccc}
\hline
\multirow{2}{*}{Source Dataset} & \multicolumn{3}{c|}{GTAv}           & \multicolumn{3}{c}{SYNTHIA}           \\ \cline{2-7} 
                                & mIOU (\%) & gain          & std(\%) & mIOU (\%) & gain            & std(\%) \\ \hline
Baseline                        & 52.1      & --            & 1.6     & 48.9      & --              & 0.3     \\ \hline
w/ random vectors               & 52.3      & 0.2$\uparrow$ & 2.3     & 48.0      & 0.9$\downarrow$ & 0.8     \\
w/ one-hot vectors              & 53.0      & 0.9$\uparrow$ & 0.8     & 49.8      & 0.9$\uparrow$   & 0.4     \\
w/ CLIP~\cite{radford2021learning}             & 54.6      & 2.5$\uparrow$ & 0.6     & \textbf{50.6 }     & 1.7$\uparrow$   & 0.5     \\
w/ word2vec~\cite{mikolov2013distributed}   &
  \textbf{55.0} &
  2.9$\uparrow$ &
  0.5 &
  50.2&
  1.3$\uparrow$ &
  0.4 \\ \hline
\end{tabular}}
\caption{The results of different vectors choices include the adaptation from GTAv and SYNTHIA}
\label{tab:vectors}
\end{table}

\subsection{Diagnostic Studies}
\label{experiments:diagnosis}

\noindent$\bullet$\quad\textbf{The choice of domain-agnostic prior.} We investigate three other options except for word2vec embedding~\cite{mikolov2013distributed}, namely, (1) that generating a $300$-dimensional, normalized \textbf{random vector} for each class, (2) directly creating a \textbf{one-hot vector} for each class, where the dimensionality equals to the number of classes, (3) classes embeddings generated by the language branch  of \textbf{CLIP}~\cite{radford2021learning}.
Similar to the main experiments, we run each option three times and report the averaged accuracy. Results are summarized in Tab~\ref{tab:vectors}.

One can see that random vectors achieve a slight $0.2\%$ accuracy gain on GTAv$\rightarrow$Cityscapes, but incurs a $0.9\%$ accuracy drop on SYNTHIA$\rightarrow$Cityscapes, implying the instability. Throughout the three individual runs on GTAv$\rightarrow$Cityscapes, the best run achieves a $54.4\%$ mIOU, just $0.6\%$ lower than using word2vec, but the worst one reports $49.2\%$ which is even significantly lower than the baseline. Regarding the one-hot vectors, the results over three runs are less diversified, and the average improvement is consistent, (\textit{i.e.}, $0.9\%$  on both GTAv$\rightarrow$Cityscapes and SYNTHIA$\rightarrow$Cityscapes), though smaller than that brought by word2vec embedding. 
The prior from CLIP~\cite{radford2021learning} reports 54.6\% and 50.6\% mIOUs on the GTAv and SYNTHIA experiments, respectively. Despite the fact that CLIP is stronger than word2vec, the mIOUs are just comparable to that using word2vec (55.0\% and 50.2\%).
From these results, we learn the lesson that (1) even a naive prior alleviates the inter-class confusion caused by domain shift, however, (2) it would be better if the inter-class relationship is better captured so that the model is aware of semantically similar classes -- text embedding offers a safe and effective option. (3) The limited number and diversity of target categories may have diminished the advantages of a stronger language model (\textit{e.g.}, CLIP), and we still pursue for a vision-aware yet domain-agnostic embedding method.

\noindent$\bullet$\quad\textbf{Different Backbones.} To verify the effectiveness of DAP on different network structures, we replace the convolution backbone (ResNet101) with a transformer network (ViT-Base~\cite{dosovitskiy2020image}). The numbers of the block layers, token size, and heads are $12, 768$ and $ 12$ respectively in the transformer encoder. The input size of the training data is set $ 768\times 768$. And we initiate the ViT encoder with a model pre-trained on ImageNet-21k then fine-tune the segmentation network with a base learning rate of $0.01$ adopted with the 'poly' learning rate decay and use SGD as the optimizer.  The $g_\mathrm{pr}$ is one layer transformer structure and weight of $\mathcal{L}_\mathrm{DAP}$ is $0.25$. On GTAv, source-only, DACS, DAP report $49.4\%$, $58.4\%$, $61.1\%$ mIOUs. As for the transferring from SYNTHIA, the numbers of the three settings are $42.7\%$, $53.2\%$, $59.1\%$ when evaluating on 16 classes and $48.1\%$, $60.9\%$, $66.1\%$ on 13 classes. We can see that DAP still obtains consistent accuracy gain. To the best of our knowledge, the DAP numbers are \textbf{SOTA}.

\vspace{0.1cm}
\noindent$\bullet$\quad\textbf{The importance of feature interpolation.} From the model that only uses the source domain, we gradually add mixed data (by DACS~\cite{tranheden2021dacs}), introduce DAP (using the word2vec embedding), and perform feature interpolation to down-sample the embedding map. As shown in Tab~\ref{tab:ablation_gta}, for the both two transferring scenarios, feature interpolation contributes nearly half accuracy gain of DAP over DACS. Intuitively, feature interpolation enables the features on small-area classes to be preserved, yet nearest-neighbor down-sampling can cause these features to be ignored.

\begin{table}[!t]
\centering
\setlength{\tabcolsep}{0.12cm}
\resizebox{0.48\textwidth}{!}{%
\begin{tabular}{c|ccccc}
\hline
Setting   & source only & DACS & DAP w/o interp & DAP w/ interp \\ \hline\hline
GTAv    & 35.5                      & 52.1 & 53.8 & \textbf{55.0}\\
SYNTHIA & 30.0   & 48.9 & 49.6 & \textbf{50.2}  \\ \hline
\end{tabular}}
\caption{Ablation on the contribution of  each module in both GTAv$\rightarrow$Cityscapes and  SYNTHIA$\rightarrow$Cityscapes experiments. }
\label{tab:ablation_gta}
\end{table}


\noindent$\bullet$\quad\textbf{Parameter Analysis.} Lastly, we study the impact of the coefficient $\alpha$ that balances between $\mathcal{L}_\mathrm{seg}$ and $\mathcal{L}_\mathrm{DAP}$. The results shown in Tab~\ref{tab:alpha} suggest that $\alpha=1.0$ is the best option. In addition, increasing $\alpha$ causes a larger accuracy drop compared to decreasing it, which may indicate that $\mathcal{L}_\mathrm{seg}$ is the essential goal and $\mathcal{L}_\mathrm{DAP}$ serves as an auxiliary term.

\begin{table}[!t]
\centering
\resizebox{0.40\textwidth}{!}{%
\begin{tabular}{c|ccccc}
\hline
$\alpha$    & 0.50 & 0.75 & 1.00          & 1.25 & 1.50 \\ \hline\hline
GTAv    & 54.5 & 54.1 & \textbf{55.0} & 54.3 & 54.7 \\
SYNTHIA & 50.0 & 49.4 & \textbf{50.2} & 49.6 & 49.1 \\ \hline
\end{tabular}}
\vspace{0.1cm}
\caption{Impact of tuning the balancing coefficient, $\alpha$. All numbers are segmentation mIOU (\%).}
\label{tab:alpha}
\end{table}



\section{Conclusions}
\label{conclusions}

In this paper, we investigate the UDA segmentation problem and observe that semantically similar classes are easily confused during the transfer procedure. We formulate this problem using the Bayesian theory and owe such confusion to the weakness of likelihood, \textit{e.g.}, insufficient training data. To alleviate the issue, we introduce domain-agnostic priors to compensate the likelihood. Experiments on two standard benchmarks for UDA segmentation in urban scenes verify its effectiveness both quantitatively and qualitatively 

\vspace{0.1cm}\noindent
\textbf{Limitations of this work.}  Currently, the best option of DAP is to leverage the text embedding vectors. We are looking forward to more powerful priors, \textit{e.g.}, from the cross-modal pre-trained models~\cite{radford2021learning,jia2021scaling}. This may call for more sophisticated designs of prior embedding, projection, alignment, \textit{etc.}, which we leave for future work.

\vspace{0.1cm}\noindent
\textbf{Acknowledge}\quad This work was supported in part by the National Natural Science Foundation of China under Contract U20A20183 and 62021001, and in part by the Youth Innovation Promotion Association CAS under Grant 2018497. It was also supported by the GPU cluster built by MCC Lab of Information Science and Technology Institution, USTC.

{\small
\bibliographystyle{ieee_fullname}
\bibliography{egbib}
}

\end{document}